\documentclass{article}
\usepackage{ijcai22}

\usepackage{times}
\usepackage{soul}
\usepackage{url}
\usepackage[hidelinks]{hyperref}
\usepackage[utf8]{inputenc}
\usepackage[small]{caption}
\usepackage{graphicx}
\usepackage{amsmath}
\usepackage{amsthm}
\usepackage{booktabs}
\usepackage{algorithm}
\usepackage{algorithmic}
\usepackage{url}

\usepackage{fancyhdr}

\fancypagestyle{firstpage}
{
    \setlength{\headheight}{11.0pt}
    \addtolength{\topmargin}{-11.0pt}
    
    \fancyhead[L]{To appear in Proceedings IJCAI 2022}  
    \fancyhead[R]{}
}

\title{Abstraction for Deep Reinforcement Learning}

\author{
Murray Shanahan$^{1,2}$
\and
Melanie Mitchell$^3$
\affiliations
$^1$DeepMind
$^2$Imperial College London
$^3$Santa Fe Institute
\emails
m.shanahan@imperial.ac.uk,
mm@santafe.edu
}

\begin{document}

\thispagestyle{firstpage}

\maketitle

\begin{abstract}
We characterise the problem of abstraction in the context of deep reinforcement learning. Various well established approaches to analogical reasoning and associative memory might be brought to bear on this issue, but they present difficulties because of the need for end-to-end differentiability. We review developments in AI and machine learning that could facilitate their adoption.
\end{abstract}

\section{Introduction}

\label{introduction}

Consider an embodied agent, either an animal (human or non-human), a physical robot, or a virtual agent in a simulated environment. Suppose the agent is exposed to a certain amount of experience from which it can learn. At every moment, the agent must address the question ``How can I shape my future experience to my liking?'' by answering the question ``How does my present experience resemble what I have experienced in the past?''\footnote{Or, to be more precise, the mechanism that underlies the agent's choice of actions must implicitly address the first question by answering the second.} How well an agent does this — that is to say how well it {\em generalises} from past experience — is one measure of its intelligence.

In contemporary deep reinforcement learning (DRL), answers to these two questions are incorporated in an agent's learned policy, a function that maps observations to agent actions (or distributions over actions), maximising cumulative discounted reward. This function is realised by a deep neural network, and generalisation from past experience is achieved through the standard tools of backpropagation and stochastic gradient descent. DRL has yielded impressive results, and must surely count as one of machine learning's most important advances of the past decade. However, in its current guise, DRL inherits the well-known shortcomings of contemporary neural network methods, which include low sample efficiency, limited transfer ability, and poor out-of-distribution generalisation \cite{garnelo2016towards,lake2017building}.

We believe these shortcomings result from an inability to form sufficiently general abstractions. However, we see no in-principle barrier to developing DRL agents that are endowed with a greater capacity for abstraction, enabling them to generalise better from past experience \cite{mitchell2021abstraction}. The purpose of this paper is to review past and present work in artificial intelligence that has the potential to advance this aim. We will discuss a number of relevant symbolic and neural-network approaches that purport to yield more abstract, more general representations. For each approach, we will make the case for its applicability, review progress, and identify the challenges it poses.

\section{Abstraction: Framing the Problem}

A human being's capacity to shape the future according to their liking via abstraction exceeds anything we can achieve in artificial intelligence today. The fundamental cognitive operation behind this ability is that of {\em analogy-making} or {\em seeing similarity}. To see a similarity is to identify certain respects in which one thing (a state or a sequence) is like another, and to conclude that they are alike in certain other respects. Seeing similarities (analogies) among a {\em set} of things is the basis for a trio of further fundamental cognitive operations: 1) forming a concept of which those things are all instances, 2) seeing that something is an instance of a concept so formed, and 3) applying the concept to infer something new about the thing in question. We will use the umbrella term {\em abstraction} to denote this cluster of operations: seeing similarity (analogy-making), forming a concept, and applying a concept.

\subsection{Common sense and transfer}

Abstraction, as conceived here, is the key to transfer. Things that appear different on a superficial level can present similarities on a deeper, more abstract, level, similarities that can be exploited in the service of transfer. To achieve this, an agent must be able to form concepts that can be usefully applied in contexts different from those in which they were acquired. In other words, the domain of a concept's application must be larger than the domain of its acquisition. The larger and more diverse the domain of a concept's application is, compared to the domain of its acquisition, the wider the {\em transfer gap} it enables an agent to cross.

The concepts with the largest and most diverse application are those belonging to ``foundational common sense’’ \cite{spelke2000core,shanahan2020artificial}. Humans and, to a lesser degree, other animals possess a repertoire of these fundamental concepts that includes, in the domain of everyday physics, such concepts as object, path, obstacle, portal, container, and so on, and, in the social domain, such concepts as other agents, being with others, meeting, giving, taking, helping, and so on. Foundational common sense concepts, though acquired through basic sensorimotor interactions with the everyday world, also structure our thinking at a more abstract level \cite{lakoff1980metaphors}. Life is a journey; every career has its obstacles; relationships are a matter of give and take.

Let's consider an example. Josie is five years old. Most weekdays, she goes to the nursery with her mother, which entails leaving the house, getting in the car, a short drive, getting out of the car, and passing through the nursery entrance. At the weekend, Josie often goes to the shops with her father, which entails leaving the house, walking up the hill and round the corner, and entering the supermarket through the sliding doors. From these and other similar experiences, Josie has distilled the concept of a journey. This concept captures a recurring pattern, a patchwork of similarities, in certain of the episodes that make up Josie's life. These episodes typically involve a point of departure, a period of motion along a path, a variety of obstacles, and a destination.

Each of the elements that features in Josie's concept of a journey can be instantiated in many ways; the place of departure could be a house or a shop, the movement might involve a car or not, the obstacles might include a door or the corner of a building. Josie's concept of a journey is indifferent to many other details too: the colour of the car, who else is in the street, and so on. Primarily Josie's concept of a journey captures narrative structure: the events involved, their types, and their order. Thanks to its concern with narrative structure and its indifference to detail, this concept can usefully be applied in new and unfamiliar circumstances. Consequently, when Josie plays a new computer game that involves retrieving a magic ring from a far off castle, she knows to look for a means of transport. Perhaps the unicorn will carry her there.

\subsection{The limitations of today's DRL agents}

A major obstacle to achieving artificial general intelligence would be overcome if we knew how to realise abstraction in DRL agents. Currently, we don't know how to endow them with the fundamental cognitive operations characterised in the previous section: seeing similarity (analogy-making), forming a concept, and applying a concept. Of course, our current deep learning systems do achieve a degree of generalisation. If a model is trained on a dataset drawn iid from some distribution, and the resulting model performs well on a separate test set that is drawn iid from the same distribution, then we can justifiably claim that it has generalised from the training set to the distribution as a whole. Moreover, there are numerous research papers describing systems that are claimed to achieve transfer, or systematic generalisation, or generalisation to held out combinations of features \cite{kirk2021survey}. However, in all cases, the type of transfer achieved is severely limited.

\begin{figure}[t]
  \centering
  \includegraphics[width=0.4\textwidth]{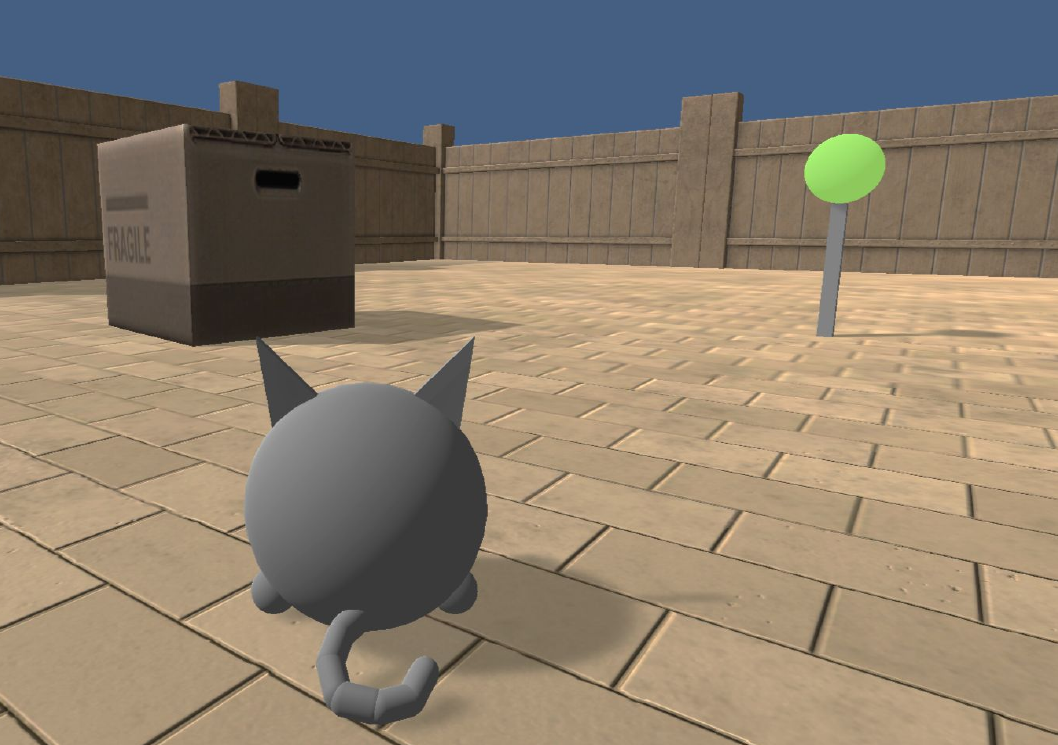}
  \caption{An example Animal-AI task. The green sphere is the reward item. The subject (human or AI) has to push the box into the pillar to knock the green sphere off.}
  \label{fig:figure}
\end{figure}

Voudouris {\it et al.} \shortcite{voudouris2022direct} explicitly compare the performance of children and agents submitted to the 2019 Animal-AI competition \cite{crosby2020animal}. In this competition, developers submitted agents for testing on a battery of tasks inspired by the animal cognition literature (Fig.~\ref{fig:figure}). Importantly, the test tasks were not revealed to the developers in advance. Instead, developers were supplied with a 3D environment to train their agents in, comprising an arena and an inventory of objects that included green and yellow spheres as rewarding ``food'' items. They were also given an indication of the general types of tasks their agents would be required to solve to obtain the food items. In the work of Voudouris {\it et al.}, subjects aged between six and 10 years were asked to solve a subset of these tasks of varying degrees of difficulty. They found that the children were able to solve all the tasks, significantly out-performing the AI agents submitted to the 2019 competition on all but the easiest levels.

Of course, the children and the AI agents differed enormously in the previous experience they brought to bear on these tasks. The children benefited from several years' experience observing and interacting with the real world in all its variety and complexity, not to mention the innate endowment of human evolutionary history. By contrast, the AI agents' only experience was of a very limited virtual world containing very few types of object (and no other agents). Nevertheless, it's instructive to consider the considerable transfer gaps the children were able to cross to solve tasks that defeated the AI agents. In one task, for example, the food item rested on a pillar too high for the subject to reach (Fig.~\ref{fig:figure}). However, it was possible to knock the food off the pillar by pushing a tall box into it. 42\% of the children (and 60\% of the ten-year olds) solved this task, whereas none of the AI agents did.

To pin down the transfer gap here, we need to know what relevant prior experience the children had. Most relevantly, they were shown a two-minute video that introduced the Animal-AI environment, and they were allowed to play a series of tutorial levels. However, neither the video nor the tutorial levels included objects on pillars. On the other hand, the children will certainly have encountered similar situations in the real world. They will have had the experience of wanting something that is out of reach. They will have had the experience of poking or pushing one object with another to make the first object move. They may or may not have faced a directly analogous situation, such as a ball stuck high in a tree but within reach of a stick. Either way, the children's success will have depended on a formidable capacity to map their previous experience onto a novel scenario in an unfamiliar virtual environment \cite{tsividis2017human,dubey2018investigating}.

\subsection{Exploiting similarity for generalisation}

\label{exploiting_similarity}

We would like to be able to emulate this capacity in a DRL agent. To see what this could mean in practice, let’s imagine how such an agent might solve the pillar task zero-shot. Pending the actual development of an agent that fits the description, this will necessarily be a speculative ``just-so'' story. But it will help us pin down the specification of the sort of architecture we are aiming for.

Suppose the agent is presented with the view we see in Fig.~\ref{fig:figure}. The raw image is encoded into a representation at the level of objects and their features and relations, which triggers the retrieval or ``activation'' of a set of concepts that match the agent's situation. Although the agent has never before had to get hold of a sphere perched on tall pillar, it has frequently attempted to obtain some target object that is not immediately to hand, either successfully or unsuccessfully, and the attempt has often involved overcoming (or trying to overcome) some difficulty in order to bring the object within its grasp. The agent has, accordingly, formed a concept that captures the similarities among all these experiences. This concept can be thought of as a {\em template} characterised by certain actors and objects, namely the agent and the target, along with landmark states and prototypical sequences of events.\footnote{Cf: Schank's concept of a {\em script} \shortcite{schank1977scripts}. Note that we are not implying that templates are structures that exist explicitly in a machine's memory. Rather, they are a way of characterising flexible, adaptive {\em behaviour}. In terms of underlying implementation, templates could be represented in a compressed and / or distributed form.}

To activate a concept, on this view, is to match a template to the situation at hand by binding (a subset of) its actors, objects, states, and events to actual actors, objects, states, and events. Some of the states and events will be in the past, and some will be in the future, suggesting possible courses of action. In this case the actors and objects are, respectively, the agent itself and the green sphere. The landmark states include the present situation with the sphere on the pillar and the agent some way away from it, and a future state where the agent is in possession of the sphere. The prototypical sequence of events will include movement towards the green sphere. But the disadvantageous position of the sphere means that this template alone isn’t enough to characterise a future that makes sense, that coheres with the agent's model of intuitive physics. When the agent arrives at the pillar, the sphere isn’t going to magically fall at its feet. So there’s a gap between the final state according to the template and the intermediate state of being next to the pillar.

To fill this gap, another concept needs to be activated. Something has to happen to knock the sphere off the pillar. But on many occasions in its past, the agent has made one object move by prodding or pushing it with another object. It has poked spheres with sticks, shoved boxes against each other, and so on. So the agent has formed a concept that distils the similarities among these experiences, the template for a ``small spatial story’’ \cite{turner1998literary}. The sole actor in this template is again the agent itself, and it features a tool object that the agent holds plus a target object to which that tool is applied. The prototypical sequence of events comprises the act of applying the tool object to the target object and the consequent movement of the target.

Binding the tool object to the box and the target object to the green sphere results in a match for this template, which in turn helps to fill the gaps in the overall template. Further concepts representing similar stereotypical narratives are activated to complete the story (the box must be retrieved and carried to the pillar). The resulting set of filled-in templates is blended together to form a coherent whole, an executable plan leading from the current state, with the sphere some distance away on the pillar, to a state with high value in terms of expected reward, with the sphere on the ground near the agent.

\section{Approaches to Abstraction}

To reiterate, the account just presented is purely speculative. We don’t yet know how to build a system capable of such abstraction, because we don't know how to realise the underlying cognitive operations of seeing similarity, concept formation, and concept application in a machine. However, the field of artificial intelligence has developed a plethora of relevant techniques. The aim of the following sections is to present an inventory of techniques with the potential to realise the cognitive operations required for abstraction, drawing on ideas from the field's past and on recent developments in machine learning.

\subsection{Structure mapping}

Gentner and collaborators proposed that the analogical-mapping process underlying abstraction is one of \textit{structure mapping}; human analogy-makers map systematic conceptual \textit{structures} rather than unconnected components \cite{gentner1983structure,falkenhainer1989structure}. The structures in question can be represented as graphs whose nodes represent {\em objects}, {\em attributes}, and {\em actions}, and whose edges represent {\em relations}. A situation (or state) encountered by an agent can be mapped to a previously encountered situation by 1) generating graphs corresponding to the two situations, and 2) mapping one graph to the other by finding a set of systematic correspondences between similar structures of nodes and edges. For example, the situation depicted in (Fig.~\ref{fig:figure}) could be represented by a graph including the sphere (food), pillar, and box, along with their attributes (e.g. colour) and spatial relationships, as well as the goal of the agent to obtain the food. A prior situation triggered in memory might include desired food on top of a fence, which the agent shook by pushing a rock into it, thus causing the sphere to drop.  This situation could be similarly represented as a graph, in which the system of objects, attributes, actions, and relations would be mapped to the current-situation graph, allowing the agent to infer that pushing the box into the pillar might analogously yield a favourable result.

To form a concept given a set of similar states or sequences is to find an intermediate graph to which they can all be mapped in this way. Then, to apply that concept to a given state is to 1) map the graph X representing the state to the graph Y representing the concept (matching), and 2) supplement X with nodes and edges (things and relations) present in Y but lacking in X (completion). A crucial feature of the matching operation is that it is not exact. There can be nodes in graph X (e.g., the colour of an object) for which there is no corresponding node in graph Y, and vice versa. Part of the task of cognition here is to learn what to ignore, what is and what is not relevant. Moreover, two nodes do not have to be identical to correspond, only to be similar. In this sense, seeing similarity is a recursive operation, which allows concepts to be organised into a hierarchy of abstraction by collapsing sub-graphs into nodes that feature in other graphs at a deeper level of abstraction.

The final inferential step (completion) is what makes a concept useful for guessing how one portion of a scene will look given the appearance of another, or for predicting the future from the past, or, more generally, for filling in the gaps when things are partially observable due to the constraints imposed on an agent by time and space. Given a repertoire of concepts, the task of cognition is to find a coherent subset of that repertoire that matches the sequence of states unfolding for an agent (retrieval), maximising the extent to which the agent can predict the future and, more specifically, can anticipate the effects of its actions. The greater the predictive power of its repertoire of concepts, the larger its spatial and temporal compass, the more the world can be said to ``make sense'' to the agent.

A significant difficulty with the structure mapping approach is the need to learn discrete representations, which are tricky to accommodate in an end-to-end differentiable processing pipeline \cite{crouse2021neural}. Moreover, in the structure mapping approach, the cognitive operations of seeing similarity, forming a concept, and applying a concept take place after the world has been carved into discrete chunks with hard boundaries. Structure mapping assumes that a distinct ``perceptual module'' is responsible for generating the representations of the current state and prior memory. However, the representation and mapping processes are likely to be strongly interdependent in any intelligent agent; the best way to represent a new situation depends on the context of the prior situations it is mapped to.

The same consideration ought to apply to a DRL agent. As Brian Cantwell Smith asserts, ``taking the world to consist of discrete intelligible mesoscale objects is an {\em achievement} of intelligence, not a premise on top of which intelligence runs''~\cite{smith2019promise}. The problem would be mitigated if the task of discovering objects and relations in raw data could be integrated in an end-to-end differentiable DRL architecture (thanks to the top-down flow of influence, via gradient-based update, from reward through action to perception) \cite{garnelo2019reconciling}. Considerable recent progress along these lines has been reported in the deep learning literature \cite{burgess2019monet,greff2019multi,engelcke2019genesis,locatello2020object,creswell2021unsupervised}, but it remains to be seen whether any of these methods are sufficiently general to support genuine abstraction.

\subsection{Other approaches to analogy}

Hofstadter {\it et al.}'s Active Symbol Architecture (ASA) for analogy-making \cite{Hofstadter1994Copycat,Hofstadter1995Fluid} shares with structure mapping the need for discrete representations, but eschews the compartmentalisation of representation and mapping. In the ASA, the construction of each situation representation is interleaved with the construction of a mapping between situations. The ASA uses an agent-based search process, inspired by earlier ``blackboard architectures'' \cite{Lesser1975Organization}, and is related to ideas from global workspace theory \cite{Baars2013Global,shanahan2006cognitive,goyal2022coordination}.

All these architectures have a common character. A set of parallel, independent, specialist processes (or ``agents'' or ``modules'') compete and co-operate to modify a shared, global memory structure (the ``workspace'' or ``blackboard''), and the state of that memory structure is, in turn, visible to the full cohort of processes. The use of a shared workspace promotes integration among the specialist processes, which must all be capable of manipulating representations in a common format. If such an architecture is realised in a neural network, the bottleneck of a shared workspace will pressure processes to collectively learn a common representational format, which in turn leads to greater abstraction and better generalisation \cite{goyal2022coordination}.

In addition to symbolic approaches such as structure mapping and ASA, several hybrid symbolic-connectionist approaches to analogy have also been explored, such as the ACME, LISA, and DORA systems \cite{Holyoak1989ACME,Hummel1996Lisa,Doumas2008DORA}. Attempts have also been made to apply deep learning directly to analogical reasoning problems \cite{Barrett2018Abstract,Zhang2019aRAVEN}. However, these deep learning models, while they can achieve high performance on specific test sets, are trained on large datasets in the domain in which the analogies are made, and the trained models have no application outside that domain. This is, in a sense, the very opposite of the kind of abstraction we are seeking, the essence of which is generic applicability.

\subsection{Associative memory}

Associative memory is a long-established area of study in machine learning. Associative memories are not usually studied in the context of abstraction, yet they meet several of our requirements. A {\em Hopfield network}, for example, combines matching, retrieval, and pattern completion in a single operation, albeit one that requires multiple iterations to achieve convergence \cite{hopfield1982neural,hopfield2007hopfield}. Perhaps something akin to a Hopfield network might serve as a basis for the sort of abstraction we are looking for. There are, for sure, many respects in which Hopfield networks are unsuitable for abstraction. However, there is an abundance of recent work in machine learning with the potential to bridge the gap between associative memory and the cognitive operations we have identified as central to abstraction.

A standard Hopfield network is an array of neurons with all-to-all recurrent connections. An attractor landscape of implicitly stored patterns is defined by the network's connection weights, which can be trained using a Hebbian learning rule. To retrieve a stored pattern, the array is initialised with an input pattern (e.g., an image), to which an update rule is repeatedly applied. This update rule minimises an energy function, ensuring that the network converges on an attractor. Under the right conditions, the attractor will correspond to the stored pattern that most closely resembles the input pattern. Because the input pattern can be partially blanked out or corrupted with noise, the retrieval process in effect performs pattern completion as well as matching.

Hopfield networks and their variants and relatives (e.g., Boltzmann machines) have been the subject of a large body of research spanning several decades. This includes a number of notable recent results. Krotov and Hopfield \shortcite{krotov2016dense} introduced a new parameterised energy function and corresponding update rule that dramatically increases the capacity of the Hopfield network while improving on the standard model's ability to store and retrieve correlated patterns. Moreover, they showed that by adjusting the update rule's parameter, the way the network functions can be smoothly varied from a distributed memory at one extreme, in which each stored pattern represents a cloud of features, to a prototype memory at the other, where each stored pattern represents a single exemplar.

Known as {\em dense associative memory} or the {\em modern Hopfield network}, this model was generalised by Demircigil {\it et al.} \shortcite{demircigil2017model}, who introduced an exponential term into the energy function. This results in much faster convergence, with a high probability of convergence after a single update step. Demircigil {\it et al.}'s energy function and update rule, which accounted for discrete-valued patterns, is further generalised by Ramsauer {\it et al.} \shortcite{ramsauer2021hopfield} to the continuous case. Ramsauer {\it et al.} also show that the resulting model is mathematically equivalent to {\em self-attention}, as used in transformer architectures \cite{vaswani2017attention}. Not only does this bring Hopfield networks, in their powerful modern guise, within the ambit of contemporary deep learning, it does so by relating them to one of the most successful innovations in the recent history of artificial intelligence.

Since they combine retrieval, matching, and pattern completion, associative memories could perhaps be used directly to see similarities between past events and present circumstances. And since modern Hopfield networks and transformer architectures are so closely related, they could potentially fulfil this role in the context of DRL. The first step would be to integrate an associative memory into a DRL architecture so that it can be used as a predictive model in the way sketched in Section \ref{exploiting_similarity}.

There have been many successful attempts to incorporate a learned predictive world model into a DRL architecture \cite{racaniere2017imagination,ha2018recurrent,gregor2019temporal,schrittwieser2020mastering}. To date, as far as we are aware, associative memories have not been used for this purpose. However, while images are the canonical use case for associative memory, (modern) Hopfield networks can be used to store and retrieve any sort of data, including sequences of images or feature vectors. In the context of sequential data, matching and retrieval can be carried out on the beginning of a sequence, with the rest of the sequence being provided through pattern completion, effectively turning an associative memory into a predictive world model. So, given its relationship to (differentiable) transformers, the modern Hopfield network could, in principle, play the part of a world model in a deep reinforcement learning architecture.

\subsection{Latent spaces}

Recall that our aim here is to endow deep reinforcement learning agents with the ability to generalise better from their past experience by seeing similarity and making analogies. In Section \ref{exploiting_similarity}, we envisaged an agent equipped with a repertoire of stereotypical templates for ``small spatial stories'', each with very wide application, such as the obtain-object template. Hopfield-style associative memories look like good candidates for realising such templates, since their attractor dynamics ensures that every situation maps to one of a small, discrete set of patterns. The obtain-object template, for example, might be represented by a corresponding stored pattern, one that would match the situation facing the agent in the pillar task, even though that pattern was learned through past experiences of situations that were very different in appearance.

However, conventional Hopfield-style associative memories, as they stand, are not up to the job. One respect in which they fall short is that they exactly reproduce specific stored patterns. This is not what we are looking for when an agent matches its current situation to a template and fills it in accordingly. When the agent confronts the green sphere on top of the pillar, for example, we don’t want it to recreate, from episodic memory, a specific past experience involving other objects. Rather, we want it to imagine a future that involves the (unfamiliar) objects in the present scene -- the pillar, the green sphere, and the box. In this respect, our requirements are for a matching and completion mechanism that is more ``constructive'', in the manner of a generative model, such as the Kanerva machine of Wu {\it et al.} \shortcite{wu2018kanerva}, which combines an associative memory with a variational autoencoder.

Variational autoencoders (VAEs) \cite{kingma2014autoencoding} and generative adversarial networks (GANs) \cite{goodfellow2014generative} learn a relatively low-dimensional representation (the latent space) of a high-dimensional distribution (over images, say). These latent encodings are obliged to exhibit a degree of abstraction because they are compressed representations of the input observations. Moreover, as shown by Higgins {\it et al.} \shortcite{higgins2016beta,higgins2018scan}, a VAE can be coaxed into learning a {\em disentangled} representation of its input distribution, wherein independent factors, such as shape, size, and colour, are represented by distinct variables. The level of abstraction in such latent encodings is modest compared to what we are seeking here. Nevertheless, they have been shown to facilitate zero-shot transfer in reinforcement learning \cite{higgins2017darla}.

The abstraction afforded by a generative model’s latent space also has the potential to improve a DRL agent's predictive world model \cite{gregor2019temporal}. Conventional world models typically operate at too concrete a level to be used in the way we have envisaged in this paper. Our hypothetical obtain-object template, for example, should be indifferent to the appearance of the object being obtained, and although it has to commit to the occurrence of certain key events, it should have little or nothing to say about their duration or about intermediate states. The world models typically used in DRL, by contrast, make predictions in observation space. They try to predict exactly how things will look to the agent in the future, on a frame by frame basis. By making predictions in latent space instead, the model described by  Gregor et al \shortcite{gregor2019temporal} exhibits a degree of both state abstraction and temporal abstraction.

The objective of minimising reconstruction error is dominant in VAE-based approaches to learning an abstract latent space. Yet an essential aspect of abstraction is the ability to ignore, or at least de-emphasise, irrelevant surface features, a requirement that is very different to reconstruction capability. Self-supervised {\em contrastive} learning offers a different approach to learning abstract latent spaces, one that is not based on reconstruction. Augmenting DRL architectures with an auxiliary contrastive prediction loss can improve performance in standard benchmarks \cite{oord2018representation,guo2020bootstrap}. A recent body of work has shown how to use contrastive losses to pre-train image encoders that are {\em general-purpose}, in the sense that they are useful for downstream tasks without further training \cite{grill2020bootstrap,chen2020simple,mitrovic2021representation}. The resulting latent encodings have also been shown to improve data efficiency in a DRL setting \cite{banino2022coberl}.

\subsection{Abstraction through language}

Our emphasis in this paper is on abstractions, such as those of foundational common sense, that are simultaneously low-level -- in the sense that they are distilled from basic sensorimotor interaction with the world -- yet highly generic. Abstractions that are built upon language are higher-level, and arguably depend on a prior capacity for sensorimotor abstraction found even in animals that lack language. Humans, however, are exposed to language from infancy, and language, in a sense, pre-packages abstraction. It presents the infant with a ready-made framework for decomposing the world into objects and relations, and for combining and re-arranging those elements into representations that reflect the compositionality of the world. As such, exposure to human language in infancy can be thought of as turbocharging the process of acquiring low-level abstractions from sensorimotor interaction. So it's natural to wonder whether the same advantage can be conferred on DRL agents, even if they are not solving language-related tasks.

A number of authors have pursued this line of thinking \cite{luketina2019survey}. For example, Andreas {\it et al.} \shortcite{andreas2018learning} show that the provision of linguistic task descriptions during training can improve performance on supervised classification in a few-shot multi-task setting, even when linguistic descriptions are not provided at evaluation time. Extending their ideas to multi-task reinforcement learning, the authors show that the provision of linguistic goal descriptions while pre-training on expert trajectories boosts performance during subsequent reinforcement learning on previously unseen tasks without linguistic descriptions. In both these settings, the trained system is evaluated on tasks that do not involve language. Nevertheless, exposure to linguistic descriptions during training enhances performance on those tasks. The authors hypothesise that language helps to mitigate over-fitting in the supervised setting, and helps to structure the agent's exploration in the reinforcement learning setting.

Further evidence that language can beneficially shape representations in reinforcement learning tasks, even when those tasks don't inherently involve language, is provided by Lampinen {\it et al.} \shortcite{lampinen2021tell}. In their work, an agent is required to select the odd one out from a set of objects. They study both 2D and 3D settings. The 3D setting is particularly demanding because the agent cannot see all the objects at once, obliging it to remember the features of previously encountered objects as it moves around. In Lampinen {\it et al.}'s  setup, the agent has an auxiliary task, in addition to maximising reward, which is to generate a reward-relevant linguistic description of the object in front of it. During training, these are compared to ground-truth descriptions, and the resulting loss helps to structure the agent's learned representations. But at test time no language is required. In both 2D and 3D settings, the authors report, agent performance is higher with the auxiliary language task than without it, and they speculate that this is because the linguistic descriptions ``highlight abstract task structure''.

Studies like these lend support to the idea that exposure to language shortcuts the challenge of learning how to divide the world into discrete objects and relations, which is a pre-requisite for acquiring generic templates of the sort we advocate here as a basis for abstraction. Perhaps language achieves this by promoting the emergence of representational structures resembling those of classical, symbolic AI. On this view, however, in contrast to the hard-coded structures of symbolic AI, the relevant structure is present in the data, partly in the world itself, and partly in the corresponding language.

Large language models offer an intriguing glimpse of another way abstraction might be achieved through language, namely with a transformer-based predictive model trained on a large corpus of sequential data. GPT-3 is capable of few-shot generalisation on tasks it was not trained on, where the tasks in question are specified by engineering suitable prompts \cite{brown2020language}. Indeed, there is some evidence that GPT-3 can solve certain analogical reasoning problems presented this way \cite{mitchell2020can}. Perhaps a large enough corpus of sequential data with the right statistical properties (eg: written human language) can pressure a large enough transformer-based model to find a general-purpose pattern completion mechanism in its parameter space.

\section{Conclusion: The Missing Pieces}

As we have seen, there is plenty of work in artificial intelligence, much of it recent, that is relevant to building deep reinforcement learning agents capable of acquiring and using abstract representations. Yet none of these advances, even in combination, amounts to a blueprint for an agent with the capabilities envisioned at the start of the paper. None of the techniques described implements anything similar to the sort of template we hypothesised as a medium for seeing similarity, making analogies, or forming concepts. The representations learned by all these techniques remain to some degree tied to the domain in which they were acquired.

By contrast, the central feature of a template is its slots or variables, which can be bound to {\em any thing}. In formal logic terms, they are `` universally quantified''. Without variables with the potential to bind to any thing, no representational structure or computational process can find open-ended application to novel situations, such as those in which the children found themselves who were so successful on the Animal-AI test suite \cite{marcus2001algebraic,martin2019predicate,greff2020binding}. Variable binding is an inherent feature of any programming language, and program induction is a promising approach to learning abstractions \cite{lazaro2019beyond,Ellis2020DreamCoder,Evans2021Sense}. However, integrating program induction with deep reinforcement learning is challenging thanks to the usual problems of reconciling discreteness and differentiability. Work on integrating binding with neural networks has a long pedigree \cite{smolensky1990tensor}, but further work on differentiable mechanisms for variable binding is critical.

However it is achieved, variable binding would also facilitate {\em compositionality}, another essential property of abstraction (and also a feature of both natural language and programming languages) \cite{higgins2018scan}. To solve unseen problems like the pillar task in the way we have proposed requires templates to be composed. Ideally, it should be possible to bind the slots in a template to other templates, leading to a form of hierarchical compositionality.

\section*{Acknowledgments}

Thanks to Richard Evans, Felix Hill, Andrew Lampinen, and the anonymous reviewers.

\bibliographystyle{named}
\bibliography{references}

\end{document}